\documentclass[letterpaper]{article} 
\usepackage[draft]{aaai2026}  
\usepackage{times}  
\usepackage{helvet}  
\usepackage{courier}  
\usepackage[hyphens]{url}  
\usepackage{graphicx} 
\usepackage{natbib}  
\usepackage{caption} 
\usepackage{booktabs} 
\usepackage{amsmath}
\usepackage{siunitx}
\DeclareSIUnit{\Thou}{K}
\DeclareSIUnit{\Mil}{M}
\DeclareSIUnit{\Bil}{B}
\DeclareSIUnit{\Tril}{T}
\DeclareSIUnit{\ms}{\milli\second}
\DeclareSIUnit{\s}{\second}

\usepackage{xcolor}
\usepackage{fontawesome5}
\graphicspath{{Figures/}}

\newcommand{\answerYes}[1]{\textcolor{blue}{#1}}

\newcommand{\answerNA}[1]{\textcolor{gray}{#1}}

\usepackage{xspace}

\newcommand\eg{e.\,g.\xspace}

\newcommand{\para}[1]{\vspace{.05in}\noindent\textbf{#1}}

\makeatletter
\renewcommand{\fps@figure}{htb}         
\renewcommand{\fps@table}{htb}         
\makeatother

\title{Infini-News: Efficiently Queryable Access to 1.3 Billion Processed Common Crawl News Articles}

\author{Ruggero Marino Lazzaroni\textsuperscript{\rm 1,$\dagger$}, Jana Lasser\textsuperscript{\rm 1}, Kirill Solovev\textsuperscript{\rm 1}}

\newcommand{\hflogo}{\raisebox{-0.2ex}{\includegraphics[height=1.4ex]{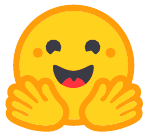}}}
\newcommand{\cblogo}{\raisebox{-0.2ex}{\includegraphics[height=1.4ex]{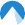}}}
\providecommand{\href}[2]{#2}
\affiliations{
	\textsuperscript{\rm 1}University of Graz, Austria \\
	\{ruggero.lazzaroni, jana.lasser, kirill.solovev\}@uni-graz.at \\
	\textsuperscript{$\dagger$}Corresponding author
	\par\smallskip
	{\footnotesize
	\hflogo~\href{https://huggingface.co/datasets/ruggsea/infini-news-corpus}{\texttt{ruggsea/infini-news-corpus}}~\,$\cdot$\,~%
	\hflogo~\href{https://huggingface.co/datasets/ruggsea/infini-news-index}{\texttt{ruggsea/infini-news-index}}~\,$\cdot$\,~%
	\cblogo~\href{https://codeberg.org/ksolovev/infini-news}{\texttt{codeberg.org/ksolovev/infini-news}}}
}

\usepackage[hidelinks]{hyperref}
\begin{document}

\maketitle

\begin{abstract}
	Large-scale news corpora support a wide range of research in Computational Social Science and NLP, yet access remains constrained: commercial archives impose prohibitive costs and licensing restrictions, while open alternatives like Common Crawl's CC-News require terabyte-scale storage and computationally intensive processing.
	We present \textsc{Infini-News}, a retrieval toolkit and index for the entire CC-News archive from August 2016 to the latest available snapshot.
	Our contributions are threefold.
	First, we extract, clean the text, and parse the structured metadata of over \qty{1.35}{\Bil} articles.
	Second, we enrich the corpus with language detection using three frontier language classifiers (GlotLID, lingua, and CommonLingua), and with multi-source geographic attribution that resolves a country of origin for 83.4\% of articles across 222 countries.
	Third, we construct Infini-gram indexes: suffix-array structures that let researchers search the full archive for arbitrary text patterns in sub-second time.
	Together, these resources lower the barrier to longitudinal, cross-national media research.
\end{abstract}

\section{Introduction}
\label{sec:introduction}

Large-scale news corpora are fundamental resources for Computational Social Science (CSS) and Natural Language Processing (NLP) \cite{Metzler2016}.
While social media data primarily captures public reactions, news archives provide a common baseline for analyzing institutional information flows, agenda-setting, and framing effects \cite{Costa2021, Weber2018, Peris2021, Liu2023, Field2018}.
Recent work has used such data to study interactions between professional journalism and online discourse during crises and political events \citep[e.g.,][]{Pierri2021, Wang2024, Pecile2025, Chen2022}.
Because social phenomena often transcend national borders, research based on media from a single country or language faces the risk of sampling bias \cite{Henrich2010, Hovy2021, Hershcovich2022}.
Access to multilingual data with broad geographic coverage is therefore critical to ensure the external validity and generalizability of these studies.

While previous research has relied heavily on news data, access at large scale remains constrained by financial and technical barriers.
Commercial vendors (e.g., Factiva, LexisNexis) provide extensive archives, but their high costs and restrictive terms prohibit data redistribution and limit reproducibility \cite{Karstens2023}.
Open alternatives, such as Common Crawl News (CC-News) \cite{CCF2016present}, offer hundreds of terabytes of raw HTML files, but transforming this raw data into research-ready corpora requires substantial engineering effort and computational resources.
This renders large-scale news data largely inaccessible to researchers without specialized infrastructure \cite{Hoffmann2022, angermaier2025schwurbelarchivgermanlanguagetelegram}.

In this work, we address this gap by introducing \textsc{Infini-News}, a processed, metadata-enriched, and queryable version of the entire CC-News archive from its inception in August 2016 onward.
This release lowers the barrier for researchers to access large-scale multilingual news data by providing three resources:
\begin{enumerate}
	\item \textbf{Processed Corpus:} We extract clean article text and structured metadata from \qty{180}{\tera\byte} of raw WARC files, yielding over \qty{1.35}{\Bil} articles, which removes the need for custom parsing infrastructure.
	\item \textbf{Metadata Enrichment:} We append language tags using three classifiers (GlotLID v3, lingua, and CommonLingua) and apply a geographic attribution cascade that resolves a country of origin for 83.4\% of articles across 222 countries.
	\item \textbf{Queryable Indexes:} We construct yearly indexes based on the Infini-gram architecture \cite{Liu2024}. These suffix-array structures enable researchers to search the full archive for arbitrary text patterns in sub-second time and construct datasets on demand.
\end{enumerate}

Our release follows the FAIR principles \cite{FORCE112020} and includes documentation, versioning policies, and a persistent DOI.
By employing a retrieval-first design, \textsc{Infini-News} enables researchers to query the corpus without downloading the underlying files, while the full archive remains available for bulk computational analysis.

\section{Background and Related Work}
\label{sec:background}

Researchers seeking large-scale news data face a choice between commercial services, curated academic datasets, and open web archives, with each option presenting trade-offs that motivate the design of \textsc{Infini-News}.

\para{Commercial Aggregators.}
Proprietary data sources such as Factiva and LexisNexis are established resources for media monitoring, but their utility for academic research is constrained by opaque terms of access and prohibitive costs \cite{Karstens2023}.
Because their coverage is largely determined by partnership agreements, it often under-represents digital-native or niche sources \cite{Gilbert2020}.
Available coverage is further restricted by commercial agreements capping retrieval volume and prohibiting redistribution of corpora between research teams \cite{FiilFlynn2022}, limiting reproducibility.

\para{Curated Academic Datasets.}
Beyond commercial aggregators, research teams have produced several manually or semi-automatically curated datasets.
The NELA-GT corpus \cite{Noerregaard2019}, for example, provides a ground-truth dataset for misinformation research but is limited to specific outlets and time-frames; task-specific corpora such as Newsroom \cite{Grusky2018} offer curated baselines for summarization but lack the publisher breadth and multilingual coverage needed to study media systems at scale (38 English-language publishers).
Platforms like Media Cloud \cite{Roberts2021} offer tooling for tracking media attention, though retrieving raw article text at scale for downstream tasks remains subject to access restrictions.
These datasets typically constrain the topics, languages, or time periods researchers can study.

\para{Open Web Archives.}
On the open web, CC-News provides a continuous, open-source alternative.
However, the raw archives are prohibitively large, with researchers relying on streaming and subsetting.
Streaming libraries \cite{Hamborg2017a, Dallabetta2024} analyze the crawl on the fly to extract articles matching a query, but sequentially downloading and parsing WARC files scales poorly across wider time windows.
Previous efforts to structure CC-News resulted in the subsets that are typically monolingual, topic-bound, or restricted to a narrow timeframe; for example, \citet{Mackenzie2020} extract English articles only from a 2{,}291-WARC slice covering Aug 2016 -- Mar 2018.
Both approaches result in a choise between longitudinal breadth and computational feasibility.

\para{Our Approach.}
We process the entirety of CC-News from its 2016 inception onward, preserving its multilingual, cross-country coverage.
We solve the retrieval bottleneck using the Infini-gram architecture \cite{Liu2024}, specifically deploying the Infini-gram mini suffix-array engine \cite{Xu2025}.
Unlike standard inverted indices, this structure supports efficient n-gram counting and search with lower storage overhead than full-text indexing.
This infrastructure allows the CSS community to define and extract targeted datasets on demand, bypassing the traditional engineering bottleneck.
To our knowledge, \textsc{Infini-News} is the first FAIR-compliant toolkit that enables efficient search and on-demand construction of targeted news datasets without requiring local storage of the full archive.

\section{Dataset Construction}
\label{sec:data_construction}

\textsc{Infini-News} adopts a retrieval-first architecture: researchers query indexes to identify relevant articles, then retrieve only the necessary documents.
By separating discovery from storage, this design reduces data transfer and storage requirements from terabytes to megabytes.
The system consists of three components: the processed corpus, the enriched metadata, and the searchable indexes.

\subsection{Acquisition and Preprocessing}
\label{sec:preprocessing}

We process all CC-News WARC files from the archive's August 2016 inception through the latest available snapshot using \texttt{fastwarc} \cite{Bevendorff2021} to stream the archives and retrieve HTML contents.
WARCs are constructed by StormCrawler \cite{stormcrawler}, which discovers and fetches articles continuously using RSS/Atom feeds and news sitemaps.\footnote{\url{https://commoncrawl.org/blog/news-dataset-available}}
Because the archive relies entirely on this feed-based discovery rather than formal inclusion criteria, its coverage structurally favors outlets that maintain machine-readable feeds.
We inherit this composition and apply no additional source filtering.

\para{Text Extraction.}
Raw HTML includes navigation elements, advertisements, and boilerplate code that obscure the article content \cite{Kohlschuetter2010}.
We apply \texttt{trafilatura} \cite{Barbaresi2021} to strip this boilerplate and extract the main article text, title, author, and publication date.
We selected \texttt{trafilatura} over alternatives like \texttt{newspaper3k} or \texttt{news-please} based on its superior performance (0.91 F1 score) on a 500-document benchmark \cite{Barbaresi2021}, its robust handling of malformed HTML, and its broad coverage of non-Latin scripts.

\para{Corpus Storage.}
Extracted records are serialized into JSONL files corresponding to each source WARC.
By retaining only research-relevant fields (cleaned text, title, date, source URL, and enriched metadata) we reduce storage requirements by approximately two orders of magnitude compared to the raw WARC archives.

\subsection{Metadata Enrichment}
\label{sec:enrichment}

We apply two annotation pipelines to support filtering and comparative research.

\para{Language Identification.}
We provide three language labels per article so downstream users can pick the classifier whose operating point fits their workload, in line with established practice for large-scale corpus creation \cite{wenzek-etal-2020-ccnet}. GlotLID v3 \cite{Kargaran2023}, the Hugging Face release of the GlotLID-M fastText classifier extending the original 1{,}665-label vocabulary to \num{2102} language-script labels and the broadest-coverage option of the three (reported macro F1 \num{0.92} for the underlying classifier on FLORES-200), is the primary label on full-article inputs and is run on every article of at least 50 characters with the top-1 label and its confidence retained; for sub-sentence and mixed-language fragments, where fastText and other long-text classifiers are known to degrade, we add lingua \cite{Stahl2024}, an n-gram statistical detector covering 75 languages and engineered for short inputs (reported ${\sim}$99\% accuracy on full sentences, ${\sim}$94\% on word pairs, ${\sim}$74\% on single words), on every non-empty text shorter than \num{1000} characters; and as a higher-precision paragraph-scale alternative we run CommonLingua \cite{Pleias2026CommonLingua}, a 2.35\,M-parameter byte-level CNN+attention model trained on 334 languages (macro F1 \num{0.79} vs.\ GlotLID v3's \num{0.67} on the CommonLID benchmark of held-out web-corpus paragraphs), on every non-empty article. The three columns are intentionally not collapsed into a single ``best'' label --- each classifier dominates at a different (coverage, input length, language family) operating point, and ISO 639-3 disagreements between them carry signal for researchers working with closely-related or mixed-language content.
Table~\ref{tab:classifier_coverage} summarises the coverage by each classifier.
CC-News language coverage substantially exceeds that of commercial aggregators such as Factiva (31 languages) \cite{DowJones} and LexisNexis (37 languages) \cite{LexisNexis}.

\begin{table}[!htbp]
	\centering
	\small
	\caption{Per-classifier language coverage on the \textsc{Infini-News} corpus.
		\emph{Codes} reports labels observed in our data over the classifier's full label space; \emph{Articles} is the article count and \emph{Share} is its fraction.}
	\label{tab:classifier_coverage}
	\begin{tabular*}{\linewidth}{@{\extracolsep{\fill}}lrrr}
		\toprule
		& Codes (encountered)     & Articles          & Share   \\
		\midrule
		GlotLID v3   & \num{1172} (\num{2102}) & \qty{1.352}{\Bil} & 99.6\%  \\
		lingua       & 75 (75)                 & \qty{343}{\Mil}   & 25.3\%  \\
		CommonLingua & 331 (334)               & \qty{1.357}{\Bil} & 100.0\% \\
		\bottomrule
	\end{tabular*}
\end{table}

\para{Geographic Attribution.}
We assign each registered domain a country of origin by combining five sources: (i)~country-code TLDs (\eg, \texttt{.de}, \texttt{.co.uk}); (ii)~an offline Wikidata cache of news-organisation entities and their headquarters; (iii)~four curated outlet lists (palewire, US-news-domains, GDELT multilingual, GDELT English); (iv)~structural HTML evidence (JSON-LD publisher addresses, imprint-page parses, footer-copyright entities, audience-locale tags); and (v)~a corpus-language rule for domains whose output is dominated by a language with a single-country mapping.
Each source yields a candidate attribution with a confidence score; the per-domain label is the highest-confidence non-conflicting candidate.
Held-out validation on 225 hosts places precision-among-resolved at 88.8\%.

\subsection{Indexes Construction}
\label{sec:index}

We construct per-year Infini-gram indexes using the suffix-array engine of \citet{Liu2024} and the more compact FM-index variant of \citet{Xu2025}, each storing corpus suffixes in compressed form and supporting occurrence counts for any pattern of length $m$ in $O(m \log n)$ time via two binary searches.
Because the engines operate directly on bytes, arbitrary substrings are supported with no tokeniser, stemming, or n-gram pre-computation, including phrases that cross language or script boundaries.
Each index maps matches to document identifiers, so researchers can retrieve articles by downloading the relevant corpus partitions or fetching specific records programmatically.
We index every article in the observation window except those with empty extracted text, and perform no de-duplication: retaining duplicates preserves the natural frequency distribution and supports direct study of content syndication and churnalism.
\citet{Penedo2024} additionally argue that de-duplication of multi-year Common Crawl dumps can hurt language-model training; we leave that choice to downstream users.

\para{Query latency.}
Table~\ref{tab:latency} reports query latency on the released indexes, measured through the \texttt{infini-gram-mini} engine over 50 exact-substring queries stratified into five frequency tiers from common stop words to rare technical phrases (10 queries per tier, 10 timed samples each).
Memory-mapped loading takes under \qty{70}{\ms} even for the full corpus and serves queries at single-digit-millisecond p50 latency across the entire \qty{1.357}{\Bil}-article index.
Pre-loading shards into RAM is impractical at the 117-shard scale and yields no measurable speed-up once the OS page cache is warm, so memory-mapped serving is the recommended configuration.
For comparison, the same queries run against the underlying parquet via \texttt{DuckDB} (SQL \texttt{LIKE}) take \qtyrange{800}{1050}{\ms} each on a single month, two orders of magnitude slower than the indexed query.

\begin{table}[!htbp]
	\centering
	\caption{Substring-query latency.
		Medians across 50 queries (10 per frequency tier) with 10 timed samples each; \emph{Load} is the one-time engine load cost.
		mmap reads from the OS page cache; RAM pre-loads every shard before serving.
		Full-corpus RAM is omitted due to RAM constraints.}
	\label{tab:latency}
	\begin{tabular*}{\linewidth}{@{\extracolsep{\fill}}lrlrr}
		\toprule
		Scope   & Shards & Mode & Load            & p50/95/99 (ms)     \\
		\midrule
		1 month & 1      & mmap & \qty{1}{\ms}    & 0.19 / 1.48 / 1.48 \\
		1 month & 1      & RAM  & \qty{12.7}{\s}  & 0.12 / 0.18 / 0.18 \\
		1 year  & 12     & mmap & \qty{19}{\ms}   & 0.63 / 2.30 / 2.30 \\
		1 year  & 12     & RAM  & \qty{141.8}{\s} & 0.62 / 1.46 / 1.46 \\
		Full    & 117    & mmap & \qty{69}{\ms}   & 6.48 / 7.73 / 7.73 \\
		\bottomrule
	\end{tabular*}
\end{table}

\section{Dataset Content and Quality}
\label{sec:data_description}
The \textsc{Infini-News} indexes cover the entire CC-News archive from August 2016 through the latest available snapshot, resulting in \qty{1.357}{\Bil} articles over approximately ten years.
The system supports queries at arbitrary temporal grains (\eg, daily or monthly) for longitudinal analysis.

\subsection{Descriptive Statistics}
\begin{figure}[!htbp]
	\centering
	\includegraphics[width=\columnwidth]{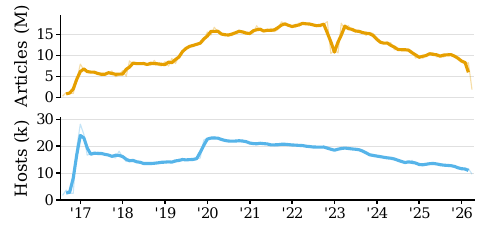}
	\caption{\textsc{Infini-News} corpus volume over time.
		Top panel (orange): articles per month, raw monthly count overlaid with a 3-month rolling mean.
		Bottom panel (blue): distinct hostnames seen per month.
		The drop in 2023 and after may be connected to the 2023 wave of \texttt{CCBot} disallow rules in news-publisher \texttt{robots.txt} files \cite{longpre2024consent}, part of a broader AI-crawler blocking trend documented by \citet{Fletcher2024}.
		The April 2026 endpoint is a partial month.}
	\label{fig:articles_and_domains_per_month}
\end{figure}

\para{Volume and Article Properties.}
The corpus contains \num{1357027742} articles with non-empty extracted text spread over 117 monthly partitions (Aug 2016 -- Apr 2026), averaging \qty{11.6}{\Mil} articles per month and peaking at \qty{18.1}{\Mil} in November 2022 (Figure~\ref{fig:articles_and_domains_per_month}).
75.1\% carry an author byline, yielding \qty{14.9}{\Mil} distinct author strings (no canonicalisation performed).
Article length is heavy-tailed, with median \num{1995} characters and per-year medians ranging from \num{1619} (2016) to \num{2044} (2026, partial).
We estimate the corpus contains \qty{1.19}{\Tril} tokens (95\% CI \qtyrange{1.190}{1.197}{\Tril}), extrapolated from a \qty{1}{\Mil}-article stratified sample tokenised with OpenAI's \texttt{tiktoken} (cl100k\_base encoding).
Mean tokens per article is 879 (median 607, mean characters/token approximately 3.4 overall but as low as 1.0 for CJK scripts).
The corpus is comparable in size to standard NLP pretraining datasets.

\para{Domain Distribution.}
Articles come from \num{133565} distinct hostnames collapsing to \num{103052} PSL-registered domains, averaging \qty{16.8}{\Thou} distinct hostnames per month; per-month variation over the corpus window is shown in Figure~\ref{fig:articles_and_domains_per_month} (bottom).
The geographic-attribution cascade resolves a country for \num{59885} of these domains (58.1\%) and for \qty{1.132}{\Bil} articles (83.4\% of the corpus) across 222 ISO 3166-1 countries.
At the domain level, the dominant attribution sources are country-code TLDs (\num{32261} domains), structural HTML and contact-page evidence (\num{11316} domains), the four curated reference lists (\num{8266} domains combined), and Wikidata including headquarters fallback (\num{7396} domains), with another 619 resolved by the article-language corpus rule.
The remaining \num{43167} unresolved domains (41.9\%) are dominated by generic-TLD outlets that lack a Wikidata entry, curated-list membership, or recoverable structural evidence.
Table~\ref{tab:source_dist} summarises the most represented countries.

\begin{table}[t]
	\centering
	\caption{Article distribution by inferred country with official country codes.
		ccTLDs are illustrative; see Section~\ref{sec:enrichment} for methodology.}
	\label{tab:source_dist}
	\begin{tabular*}{\linewidth}{@{\extracolsep{\fill}}llrr}
		\toprule
		Country                        & ccTLD        & Articles          & \%   \\
		\midrule
		United States                  & \texttt{.us} & \qty{301.3}{\Mil} & 22.2 \\
		Germany                        & \texttt{.de} & \qty{74.7}{\Mil}  & 5.5  \\
		Italy                          & \texttt{.it} & \qty{60.5}{\Mil}  & 4.5  \\
		France                         & \texttt{.fr} & \qty{55.6}{\Mil}  & 4.1  \\
		India                          & \texttt{.in} & \qty{52.2}{\Mil}  & 3.9  \\
		United Kingdom                 & \texttt{.uk} & \qty{49.7}{\Mil}  & 3.7  \\
		Russian Federation             & \texttt{.ru} & \qty{48.1}{\Mil}  & 3.5  \\
		Spain                          & \texttt{.es} & \qty{38.9}{\Mil}  & 2.9  \\
		Türkiye                        & \texttt{.tr} & \qty{38.7}{\Mil}  & 2.9  \\
		Argentina                      & \texttt{.ar} & \qty{24.0}{\Mil}  & 1.8  \\
		Greece                         & \texttt{.gr} & \qty{21.2}{\Mil}  & 1.6  \\
		Ukraine                        & \texttt{.ua} & \qty{19.9}{\Mil}  & 1.5  \\
		Mexico                         & \texttt{.mx} & \qty{17.9}{\Mil}  & 1.3  \\
		Canada                         & \texttt{.ca} & \qty{15.9}{\Mil}  & 1.2  \\
		Romania                        & \texttt{.ro} & \qty{14.0}{\Mil}  & 1.0  \\
		\midrule
		\textit{Other (207 countries)} &              & \qty{299.1}{\Mil} & 22.0 \\
		\midrule
		\textit{Unattributed}          &              & \qty{225.5}{\Mil} & 16.6 \\\bottomrule
	\end{tabular*}
\end{table}

\para{Language Diversity.}
GlotLID v3 assigns a label to \qty{1.352}{\Bil} articles (99.6\% of the corpus), spanning \num{1172} distinct ISO 639-3 codes; 166 of those codes occur in at least \num{10000} articles and 280 in at least \num{1000}.
CommonLingua scores every non-empty article across a 334-label vocabulary, of which 331 codes appear in our data; the difference from GlotLID's coverage falls almost entirely on the sub-50-character band that GlotLID's gate excludes.
The two classifiers agree on the high-resource head and most low-resource codes; on a few major languages they pick different ISO 639-3 codes for the same language (Chinese: \texttt{cmn} from GlotLID vs.\ \texttt{zho} from CommonLingua; Norwegian: \texttt{nob} vs.\ \texttt{nor}; Estonian: \texttt{ekk} vs.\ \texttt{est}).
For low-resource languages CommonLingua tends to label more articles than GlotLID, particularly in the African and Indo-Aryan tail (\eg, Eastern Mari \texttt{mhr}: GlotLID \num{6612} vs.\ CommonLingua \num{66556}; Javanese \texttt{jav}: \num{8035} vs.\ \num{49112}).
Table~\ref{tab:language_distribution} reports counts from both classifiers for selected high-resource and low-resource languages.
\begin{table}[!htbp]
	\centering
	\footnotesize
	\renewcommand{\arraystretch}{0.85}
	\setlength{\tabcolsep}{3pt}
	\caption{Language distribution in \textsc{Infini-News}.
		Article counts from GlotLID v3 and CommonLingua, each scoring articles independently; top 15 high-resource languages (by GlotLID count) followed by a selection of low-resource codes.}
	\begin{tabular*}{\linewidth}{@{\extracolsep{\fill}}llrr}
		\toprule
		Language           & Code      & GlotLID           & CommonLingua      \\
		\midrule
		\multicolumn{4}{c}{\textit{Top 15 Languages}}                          \\
		\midrule
		English            & eng       & \qty{506.8}{\Mil} & \qty{521.3}{\Mil} \\
		Spanish            & spa       & \qty{134.8}{\Mil} & \qty{135.2}{\Mil} \\
		Russian            & rus       & \qty{88.3}{\Mil}  & \qty{88.8}{\Mil}  \\
		German             & deu       & \qty{84.5}{\Mil}  & \qty{86.7}{\Mil}  \\
		Italian            & ita       & \qty{70.5}{\Mil}  & \qty{70.8}{\Mil}  \\
		French             & fra       & \qty{54.7}{\Mil}  & \qty{55.3}{\Mil}  \\
		Turkish            & tur       & \qty{44.1}{\Mil}  & \qty{43.9}{\Mil}  \\
		Arabic (MSA)       & arb       & \qty{32.1}{\Mil}  & \qty{28.9}{\Mil}  \\
		Portuguese         & por       & \qty{27.4}{\Mil}  & \qty{27.2}{\Mil}  \\
		Hindi              & hin       & \qty{20.1}{\Mil}  & \qty{21.5}{\Mil}  \\
		Greek              & ell       & \qty{18.5}{\Mil}  & \qty{18.3}{\Mil}  \\
		Japanese           & jpn       & \qty{18.2}{\Mil}  & \qty{18.4}{\Mil}  \\
		Romanian           & ron       & \qty{17.0}{\Mil}  & \qty{16.8}{\Mil}  \\
		Polish             & pol       & \qty{15.7}{\Mil}  & \qty{15.6}{\Mil}  \\
		Chinese (Mandarin) & cmn / zho & \qty{15.6}{\Mil}  & \qty{15.8}{\Mil}  \\
		\midrule
		\multicolumn{4}{c}{\textit{Selected Low-Resource Languages}}           \\
		\midrule
		Somali             & som       & \num{273506}      & \num{265535}      \\
		Yoruba             & yor       & \num{41548}       & \num{26036}       \\
		Tibetan            & bod       & \num{17384}       & \num{19030}       \\
		Sakha (Yakut)      & sah       & \num{10150}       & \num{35115}       \\
		Irish              & gle       & \num{9065}        & \num{9741}        \\
		Javanese           & jav       & \num{8035}        & \num{49112}       \\
		Eastern Mari       & mhr       & \num{6612}        & \num{66556}       \\
		Dhivehi            & div       & \num{3249}        & \num{3320}        \\
		\midrule
		Total labeled      &           & \qty{1.352}{\Bil} & \qty{1.357}{\Bil} \\
		Distinct codes     &           & \num{1172}        & 331               \\
		\bottomrule
	\end{tabular*}
	\label{tab:language_distribution}
\end{table}

Appendix~\ref{app:per_lang_temporal} plots monthly volume per language.
The high-resource head (English, Spanish, Russian, German, Italian) remains stable at millions of articles per month; low-resource codes sit four to six orders of magnitude below and are far more volatile.

\subsection{Coverage Validation}
\para{Volumetric Coverage.}
To assess volumetric coverage, we benchmark article counts against Factiva \cite{DowJones}, a widely used commercial news aggregator, restricted to high-resource languages since Factiva does not index low-resource ones.

\begin{figure}[h]
	\centering
	\includegraphics[width=\columnwidth]{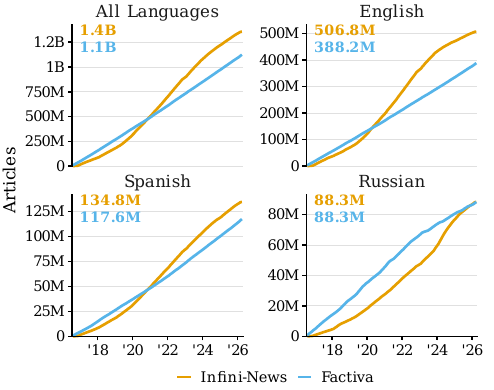}
	\caption{Cumulative monthly article counts for \textsc{Infini-News} (orange) and Factiva (blue) since August 2016.}
	\label{fig:factiva_comparison}
\end{figure}

Figure~\ref{fig:factiva_comparison} plots cumulative monthly article counts from August 2016 through April 2026 for the global total and three high-resource languages, with both series clipped to the same window.
Across the global total \textsc{Infini-News} contains roughly \qty{1.36}{\Bil} articles, against Factiva's \qty{1.12}{\Bil}; the lead is larger for English (\qty{507}{\Mil} vs.\ \qty{388}{\Mil}) and Spanish (\qty{135}{\Mil} vs.\ \qty{118}{\Mil}).
For Russian, both indexes reach ${\sim}\,$\qty{88}{\Mil} articles, with \textsc{Infini-News} narrowly ahead by the end of the window.
While the open-access CC-News pipeline matches or exceeds Factiva's volume on the high-resource end, it also covers hundreds of low-resource languages that commercial offerings lack.

\para{Comparison with fundus.}
We additionally compare \textsc{Infini-News} against fundus \cite{Dallabetta2024}, an open-source news scraper with hand-curated extraction rules for 172 publishers across 39 country codes; 154 of those 172 (89.5\%) appear somewhere in our corpus.
Because fundus and \textsc{Infini-News} both read the same CC-News WARC files, the comparison measures what each surfaces for the same query in the same window, not who can access the data.
Table~\ref{tab:fundus_comparison} reports per-query article counts over a 48-hour CC-News slice (2022-06-15 to 2022-06-17) under three configurations: \textsc{Infini-News} over the full corpus, \textsc{Infini-News} restricted to fundus' supported publishers, and fundus' real \texttt{CCNewsCrawler} output.
\textsc{Infini-News} restricted to the same hostnames returns about 30\% more articles than fundus extracts (median ratio 1.31, range 1.12--1.52), a gap dominated by fundus' stricter extraction filters.
Lifting the publisher restriction, \textsc{Infini-News} returns roughly an order of magnitude more articles than the fundus-allow-list subset, because most CC-News content falls outside fundus' 172 hand-picked outlets.

\begin{table}[!htbp]
	\centering
	\small
	\caption{Per-query article counts over the 48-hour CC-News slice 2022-06-15 to 2022-06-17.
		\emph{Fundus subset} restricts \textsc{Infini-News} to fundus' 154 supported publishers.}
	\label{tab:fundus_comparison}
	\begin{tabular*}{\linewidth}{@{\extracolsep{\fill}}lrrr}
		\toprule
		Query          & Full corpus & Fundus subset & fundus     \\
		\midrule
		Covid          & \num{86453} & \num{2695}    & \num{2061} \\
		inflation      & \num{37189} & \num{1810}    & \num{1610} \\
		Macron         & \num{13119} & 960           & 654        \\
		climate change & \num{10874} & 413           & 356        \\
		Bitcoin        & \num{10555} & 241           & 188        \\
		World Cup      & \num{8289}  & 716           & 484        \\
		Vladimir Putin & \num{8080}  & 477           & 380        \\
		Bundesliga     & \num{2432}  & 409           & 269        \\
		Roe v Wade     & 116         & 36            & 27         \\
		\bottomrule
	\end{tabular*}
\end{table}

\para{Popularity Coverage.}
To check that the corpus covers both popular and niche outlets, we compare its domains against the Tranco list \cite{LePochat2019}, which aggregates the major commercial top-sites rankings; the snapshot used here (list \texttt{6G8PX}) combines Cisco Umbrella, Cloudflare Radar, and Majestic.
Of the \num{103052} PSL-registered domains in \textsc{Infini-News}, 458 fall in Tranco's global top-1k, \num{34250} (33.2\%) in the top-1M, and \num{68801} (66.8\%) below it (Table~\ref{tab:tranco_coverage}).
Each bucket is 4--9 times the previous, the geometric growth characteristic of a Zipfian long-tail distribution.
Of the \num{68801} domains below the top-1M, \num{36223} still resolve to a country through our attribution cascade, confirming they are real publishers below Tranco's traffic cutoff.
We pin the Tranco snapshot \texttt{6G8PX} (2026-05-14) for reproducibility.

\begin{table}[h]
	\centering
	\caption{\textsc{Infini-News} registered domains by Tranco rank bucket (list \texttt{6G8PX}, 2026-05-14).
		Buckets are disjoint and span four orders of magnitude.}
	\label{tab:tranco_coverage}
	\begin{tabular*}{\linewidth}{@{\extracolsep{\fill}}lrr}
		\toprule
		Tranco rank   & Domains      & \%    \\
		\midrule
		1--1k         & 458          & 0.4   \\
		1k--10k       & \num{2307}   & 2.2   \\
		10k--100k     & \num{9754}   & 9.5   \\
		100k--1M      & \num{21731}  & 21.1  \\
		not in top 1M & \num{68801}  & 66.8  \\
		\midrule
		Total         & \num{103052} & 100.0 \\
		\bottomrule
	\end{tabular*}
\end{table}

\section{Applications}
\label{sec:use_cases}

\subsection{On-Demand Corpus Construction}
Building a longitudinal corpus around a specific event typically requires setting up custom crawlers, paying for proprietary database access, or processing raw archival dumps.
\textsc{Infini-News} replaces this with on-demand sub-corpora: a researcher filters by keyword, time range, domain, or language, and extracts the matching subset directly from the indexes.
Because each match resolves to a stable record identifier, reproducible corpus definitions can be shared as ID lists rather than redistributed as text, side-stepping the storage and licensing problems of traditional data sharing.
As a concrete example, researchers interested in the Russo-Ukrainian war could get the entire corpus of \texttt{.ru} and \texttt{.ua} articles mentioning a specific propaganda phrase since 2022, returned in seconds.

\subsection{Longitudinal Analysis}
Following longitudinal collections such as TeleScope \cite{Gangopadhyay2025}, \textsc{Infini-News} supports the study of how online news discourse changes across the entire ten-year window.
Each article carries a daily date stamp, so researchers can aggregate at any temporal grain to model story lifecycles or study specific events.
Because the corpus is built from archival WARC snapshots captured near the time of publication, records remain stable, unlike ad-hoc post-event crawls that hit link decay or proprietary aggregators that drop outlets on contract expiry; this stability speaks to broader concerns about the stewardship of online behavioral data \cite{doi:10.1073/pnas.2025764118}.

\subsection{Content Flow Analysis}
Because the Infini-gram engine supports arbitrary-length n-gram search and we deliberately preserve duplicate content, the toolkit can be used to analyse how content propagates across news sources.

\para{Coordinated Information Operations.}
Identical text appearing across multiple outlets at the same time is a marker of coordinated inauthentic behaviour.
Tracking verbatim string occurrences across domains over time lets researchers study cross-outlet reuse and narrative diffusion.

\para{Content Syndication.}
The same approach reveals legitimate syndication patterns.
Tracking verbatim article propagation across outlets quantifies wire-service penetration in national press cycles and isolates pure-republisher domains; aggregated over time, the same data exposes structural shifts in media ownership and attention concentration.

\para{Cross-Platform Information Tracing.}
Social media research often needs to relate professional news output to platform-specific content.
\textsc{Infini-News} complements social media datasets such as \textit{News on TikTok} \cite{Mayer2025} and \textit{UKElectionNarratives} \cite{Haouari2025}: scholars can query our index to find the source articles behind viral short-form videos or forwarded Telegram messages, and trace how information is reframed, moralised, or distorted as it moves from the press into social platforms.
Conversely, researchers studying the provenance of viral claims can trace them back to their original news sources.
Questions such as ``where did this statistic first appear?'' or ``which outlet first published this quote?'' become tractable, directly enabling misinformation source-tracing and fact-checking workflows.

\subsection{Comparative and Framing Research}
\para{Affective and Moral Framing.}
The dataset supports large-scale analysis of emotional and moral language in news coverage.
Building on findings about moralised language and information diffusion \citep[e.g.,][]{Solovev2023}, researchers can examine how anger, fear, and moral outrage in headlines correlate with article propagation across platforms.
Filtering by source domain enables comparative studies across outlets with different editorial orientations or factual-reliability ratings.

\para{Cross-National and Cross-Lingual Research.}
Combining language identification and geographic attribution enables systematic comparison across media systems.
Researchers can examine how national presses covered shared crises, such as the 2016 U.S.~election cycle, Brexit, COVID-19, the Russian invasion of Ukraine, and subsequent national elections, or test whether findings from one linguistic context generalise to others.
Coverage beyond the usual high-resource head supports both NLP work (developing and evaluating models for underrepresented languages) and substantive comparative research extending theories developed on Western media to broader contexts.

\subsection{Large Language Model Research}
The corpus also supports several lines of LLM research.
The indexes enable two evaluation modes: training-data contamination checks (does specific time-sensitive content appear verbatim in an LLM's training corpus?) and time-aware benchmarks that partition queries by publication date around a model's knowledge cutoff.
The \num{1172} ISO 639-3 codes that GlotLID resolves, 166 with at least \num{10000} articles each, make \textsc{Infini-News} a stress test for the multilingual coverage of frontier models on languages omitted from common evaluation suites.
The toolkit also serves as a retrieval pipeline for Retrieval-Augmented Generation (RAG) systems with article-level ground truth and source attribution.

\section{Data Access and FAIR Principles} \label{sec:data_access}

We publicly release the \textsc{Infini-News} indexes and retrieval pipeline adhering to FAIR principles.\footnote{Corpus: \url{https://huggingface.co/datasets/ruggsea/infini-news-corpus}; Indexes: \url{https://huggingface.co/datasets/ruggsea/infini-news-index}; Code: \url{https://codeberg.org/ksolovev/infini-news}}

\para{Findable.}
The corpus and indexes are hosted as separate Hugging Face datasets, each assigned a persistent DOI and accompanied by a dataset card describing coverage, schema, and versioning.

\para{Accessible.}
The corpus is available for direct download, gated to require institutional affiliation and stated research purpose.
This restriction is compatible with FAIR principles \cite{FORCE112020}, which permit authentication where necessary.
The indexes support two access modes: (1) local deployment via downloadable shards, or (2) remote querying through a hosted API.
The indexes dataset card provides documentation and code examples for API-based retrieval, enabling researchers to search and fetch articles without downloading the full corpus.

\para{Interoperable.}
The corpus is distributed in Apache Parquet with a documented per-column schema; identifiers use standard vocabularies (ISO 639-3 language codes, ISO 15924 script codes, the IPTC top-17 Media Topic taxonomy, and the original WARC URNs).

\para{Reusable.}
The retrieval toolkit's code is released under the MIT License while indexes metadata under CC-BY 4.0.
The underlying article text remains subject to original publishers' copyright; as with other web-derived corpora, the copyright status of individual documents is neither uniform nor fully auditable \cite{dodge-etal-2021-documenting, longpre2024consent}.
We provide gated access requiring institutional affiliation and stated research purpose, intended to operate within research exceptions recognized in most jurisdictions.
Downstream users bear responsibility for compliance with applicable copyright law.

\section{Limitations} \label{sec:limitations}

\begin{figure}[h]
	\centering
	\includegraphics[width=\columnwidth]{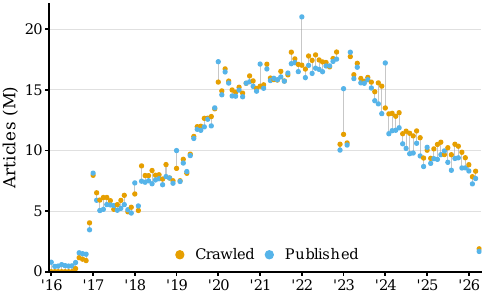}
	\caption{Per-month barbell plot of crawled (orange, \texttt{warc\_date}) and published (\texttt{blue, publish\_date}) articles.
		Most are crawled within the publication month.}
	\label{fig:crawl_publish_lag}
\end{figure}

\para{Real Time Coverage and Temporal Latency.}
Because CC-News crawls web pages with a variable delay after publication, the corpus is suited for retrospective analysis rather than real-time media monitoring.
Additionally, as a scrape of the open web, it inherently lacks paywalled content from premium outlets.
Figure~\ref{fig:crawl_publish_lag} plots the per-month count of articles by crawl date versus publication date.
The two series track closely, with month-level mismatches typically below one million articles.
Researchers studying breaking news dynamics should consider this latency, and the corpus should not be treated as a complete census of all news published during the observation period.

\para{Crawler Opt-Outs.}
The corpus exhibits a noticeable drop in monthly article volume beginning around mid-2023 (Figure~\ref{fig:articles_and_domains_per_month}).
This timing coincides with a wave of \texttt{CCBot} disallow rules added to news-publisher \texttt{robots.txt} files.
The shift was likely triggered by OpenAI's \texttt{GPTBot} announcement (7~August~2023) and Google's \texttt{Google-Extended} announcement (28~September~2023), which prompted publishers to broadly revisit their crawler policies.
\citet{longpre2024consent} document the disallow wave on \texttt{CCBot} and related crawlers across the AI training-data commons; \citet{Fletcher2024} document the broader AI-crawler blocking trend in news publishers (focused on \texttt{GPTBot} and \texttt{Google-Extended}) that publishers extended to \texttt{CCBot} in the same period.
While we do not claim a strict causal mechanism without measuring per-domain crawl status against article counts, researchers analyzing content from late 2023 onward should be aware that the publisher composition may skew away from outlets that block CCBot.

\para{Metadata Quality.}
Metadata enrichment relies on heuristics and model inferences that are subject to error.
The country-attribution cascade has three primary failure modes: country-code TLDs occasionally host content targeting other regions, structural HTML evidence reflects what the publisher declares rather than ground truth, and 41.9\% of registered domains receive no attribution at all (predominantly generic-TLD outlets lacking a Wikidata entry, curated-list membership, or structural HTML evidence).
Among resolved domains, validation against a held-out set of 225 hosts places overall precision at 88.8\%, but per-source precision varies; the imprint-VAT rule, in particular, drops to 25\% on validation, driven by cross-border VAT registrations such as Polish news organisations operating under Belgian VAT.
Similarly, while GlotLID performs well on longer texts, it may misclassify short articles, mixed-language content, or closely related languages.
The lingua short-text column mitigates but does not eliminate this issue for inputs below \num{1000} characters, and the two classifiers can disagree.
Consequently, researchers requiring precise geographic metadata or working with specific low-resource languages should independently validate the country labels and language assignments for their subset.

\para{Scope.}
The corpus only contains crawlable articles that have been published online: no comments, engagement metrics, or reader behaviour.
Research questions requiring audience-side data must draw on complementary sources.
CC-News itself only reaches back to August 2016; older content falls outside scope.

\section{Ethical Considerations} \label{sec:ethics}

\para{Consent and Privacy.}
\textsc{Infini-News} indexes publicly published news articles rather than user-contributed content.
No direct consent was obtained from individuals mentioned in coverage, as they are subjects of published journalism operating under standard editorial and legal norms.
The underlying web crawl was performed by Common Crawl following robots.txt conventions.
While the content is public, aggregating it at scale shifts what is practically accessible; gated release and usage restrictions mitigate the corresponding risks.

\para{Potential Risks.}
We identify two primary risk categories.
First, the corpus could be used to train language models without appropriate attribution or in violation of the original publishers' terms.
Second, the toolkit's capacity to detect content syndication and coordinated messaging presents a dual-use concern: while designed to support research on propaganda and media manipulation, the same capabilities could inform adversarial actors developing counter-measures against detection.

\para{Mitigations.}
We have implemented several safeguards.
The pre-processed corpus is released through a gated repository that requires institutional affiliation and a stated research purpose, while the public API returns index pointers rather than raw text, so users must obtain corpus access before retrieving article content.
Researchers granted access commit to academic use only; derived-metadata columns are released under CC-BY-4.0.
While these measures cannot eliminate misuse, they establish accountability and raise the barrier to unauthorized applications.

\para{Responsible Use.}
We encourage researchers to consider how their analyses might affect the outlets and individuals represented in the corpus.
Findings about specific sources should be communicated with appropriate context, recognizing that research outputs can influence reputations and that aggregate patterns may not reflect the practices of individual journalists or editors.

\section{Conclusion and Future Work}
\label{sec:conclusion}
We introduced \textsc{Infini-News}, a retrieval-first toolkit for the Common Crawl News archive from August 2016 through the latest available snapshot.
The suffix-array index supports n-gram search over terabyte-scale data without local storage of the full corpus; the cleaned, metadata-enriched dataset is available for researchers who need direct access for bulk computational work.
We plan to continue developing \textsc{Infini-News} along three directions:

\para{Semantic Enrichment.} We plan to gradually add pre-computed metadata layers: toxicity scores, sentiment indicators, and topic annotations.
These will let researchers filter by high-level concepts in addition to lexical patterns, broadening the framing and polarization analyses the toolkit supports.

\para{Interactive Interface.} We are developing a web-based query interface that will allow researchers to search the dataset, visualize temporal distributions, and export filtered subsets via a user-friendly front-end, reducing the technical overhead for entry.

\para{Continuous Updates.} The release covers the archive from its 2016 inception.
We commit to an annual update cycle for at least the next five years that incorporates new CC-News snapshots, keeping the longitudinal record current.

\bibliography{literature}

\section*{Paper Checklist}

\subsection{1.
	For most authors...}
\begin{enumerate}
	\item[(a)]  Would answering this research question advance science without violating social contracts, such as violating privacy norms, perpetuating unfair profiling, exacerbating the socio-economic divide, or implying disrespect to societies or cultures?
	      \answerYes{Yes.}
	\item[(b)] Do your main claims in the abstract and introduction accurately reflect the paper's contributions and scope?
	      \answerYes{Yes.}
	\item[(c)] Do you clarify how the proposed methodological approach is appropriate for the claims made?
	      \answerYes{Yes.}
	\item[(d)] Do you clarify what are possible artifacts in the data used, given population-specific distributions?
	      \answerYes{Yes.
		      Section~\ref{sec:enrichment} notes that the per-domain geographic cascade leaves 41.9\% of registered domains unresolved (predominantly generic-TLD outlets) and that language identification may underperform on low-resource languages.
		      Section~\ref{sec:data_description} clarifies that content redundancy is preserved, meaning article counts reflect propagation rather than unique documents.}
	\item[(e)] Did you describe the limitations of your work?
	      \answerYes{Yes.}
	\item[(f)] Did you discuss any potential negative societal impacts of your work?
	      \answerYes{Yes.}
	\item[(g)] Did you discuss any potential misuse of your work?
	      \answerYes{Yes.}
	\item[(h)] Did you describe steps taken to prevent or mitigate potential negative outcomes of the research, such as data and model documentation, data anonymization, responsible release, access control, and the reproducibility of findings?
	      \answerYes{Yes.}
	\item[(i)] Have you read the ethics review guidelines and ensured that your paper conforms to them?
	      \answerYes{Yes.}
\end{enumerate}

\subsection{2.
	Additionally, if your study involves hypotheses testing...}
\begin{enumerate}
	\item[(a)] Did you clearly state the assumptions underlying all theoretical results?
	      \answerNA{NA.}
	\item[(b)] Have you provided justifications for all theoretical results?
	      \answerNA{NA.}
	\item[(c)] Did you discuss competing hypotheses or theories that might challenge or complement your theoretical results?
	      \answerNA{NA.}
	\item[(d)] Have you considered alternative mechanisms or explanations that might account for the same outcomes observed in your study?
	      \answerNA{NA.}
	\item[(e)] Did you address potential biases or limitations in your theoretical framework?
	      \answerNA{NA.}
	\item[(f)] Have you related your theoretical results to the existing literature in social science?
	      \answerNA{NA.}
	\item[(g)] Did you discuss the implications of your theoretical results for policy, practice, or further research in the social science domain?
	      \answerNA{NA.}
\end{enumerate}

\subsection{3.
	Additionally, if you are including theoretical proofs...}
\begin{enumerate}
	\item[(a)] Did you state the full set of assumptions of all theoretical results?
	      \answerNA{NA.}
	\item[(b)] Did you include complete proofs of all theoretical results?
	      \answerNA{NA.}
\end{enumerate}

\subsection{4.
	Additionally, if you ran machine learning experiments...}
\begin{enumerate}
	\item[(a)] Did you include the code, data, and instructions needed to reproduce the main experimental results (either in the supplemental material or as a URL)?
	      \answerNA{NA.}
	\item[(b)] Did you specify all the training details (e.g., data splits, hyperparameters, how they were chosen)?
	      \answerNA{NA.}
	\item[(c)] Did you report error bars (e.g., with respect to the random seed after running experiments multiple times)?
	      \answerNA{NA.}
	\item[(d)] Did you include the total amount of compute and the type of resources used (e.g., type of GPUs, internal cluster, or cloud provider)?
	      \answerNA{NA.}
	\item[(e)] Do you justify how the proposed evaluation is sufficient and appropriate to the claims made?
	      \answerNA{NA.}
	\item[(f)] Do you discuss what is ``the cost`` of misclassification and fault (in)tolerance?
	      \answerNA{NA.}
\end{enumerate}

\subsection{5.
	Additionally, if you are using existing assets (e.g., code, data, models) or curating/releasing new assets, \textbf{without compromising anonymity}...}
\begin{enumerate}
	\item[(a)] If your work uses existing assets, did you cite the creators?
	      \answerYes{Yes.}
	\item[(b)] Did you mention the license of the assets?
	      \answerYes{Yes.
		      Our code is MIT-licensed; index metadata is CC-BY 4.0.
		      The underlying news content remains subject to original publishers' copyright, therefore we provide gated access for non-commercial research and note that users bear responsibility for jurisdictional compliance.}
	\item[(c)] Did you include any new assets in the supplemental material or as a URL?
	      \answerYes{Yes.}
	\item[(d)] Did you discuss whether and how consent was obtained from people whose data you're using/curating?
	      \answerYes{Yes.}
	\item[(e)] Did you discuss whether the data you are using/curating contains personally identifiable information or offensive content?
	      \answerYes{Yes.
		      It is covered in the Datasheet.}
	\item[(f)] If you are curating or releasing new datasets, did you discuss how you intend to make your datasets FAIR (see \cite{FORCE112020})?
	      \answerYes{Yes.}
	\item[(g)] If you are curating or releasing new datasets, did you create a Datasheet for the Dataset (see \cite{Gebru2021})?
	      \answerYes{Yes.}
\end{enumerate}

\subsection{6.
	Additionally, if you used crowdsourcing or conducted research with human subjects, \textbf{without compromising anonymity}...}
\begin{enumerate}
	\item[(a)] Did you include the full text of instructions given to participants and screenshots?
	      \answerNA{NA.}
	\item[(b)] Did you describe any potential participant risks, with mentions of Institutional Review Board (IRB) approvals?
	      \answerNA{NA.}
	\item[(c)] Did you include the estimated hourly wage paid to participants and the total amount spent on participant compensation?
	      \answerNA{NA.}
	\item[(d)] Did you discuss how data is stored, shared, and deidentified?
	      \answerNA{NA.}
\end{enumerate}


\appendix

\renewcommand{\thefigure}{S\arabic{figure}}
\setcounter{figure}{0}

\section*{Supplementary Materials}

\section{Per-Language Monthly Volume}\label{app:per_lang_temporal}

\begin{figure*}[t]
	\centering
	\includegraphics[width=\textwidth]{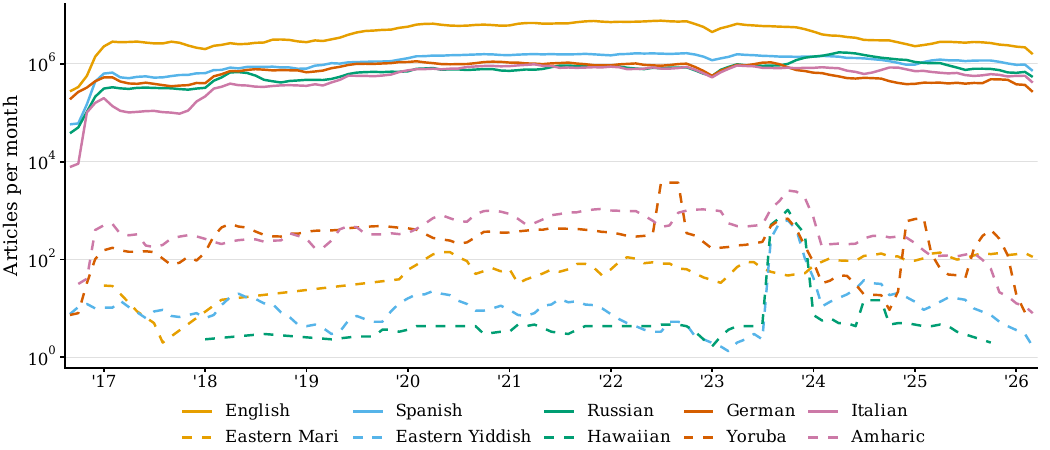}
	\caption{Articles per month for the top-5 languages by article count in \textsc{Infini-News} (solid: English, Spanish, Russian, German, Italian) and five selected low-resource codes (dashed: Eastern Mari, Eastern Yiddish, Hawaiian, Yoruba, Amharic).
		Log $y$-axis; 3-month rolling mean.
		The two groups sit four to six orders of magnitude apart in monthly volume.
		The low-resource codes shown here partially overlap with the low-resource sample in Table~\ref{tab:language_distribution} but are not identical, as the per-language temporal subset was selected to span a range of script families and geographies.}
	\label{fig:articles_by_lang}
\end{figure*}

The volumetric headline numbers in Table~\ref{tab:language_distribution} compress ten years of crawl history into a single integer per language, hiding the shape of the per-month curve.
Figure~\ref{fig:articles_by_lang} unpacks that shape on a logarithmic vertical axis, juxtaposing the five highest-volume languages with five low-resource codes drawn from across script families and geographies.
The high-resource head, plotted with solid lines, settles into a roughly stationary plateau after the 2017 seed-list expansion, with English, Spanish, Russian, German, and Italian each accumulating millions of articles per month and tracking the corpus-wide growth and 2023 contraction described in the main paper.
The low-resource codes, plotted dashed, instead sit four to six orders of magnitude lower, exhibiting the bursty, sporadic pattern typical of long-tail languages on the open web: months in which CC-News encounters a single eligible publisher alternate with months in which none surface at all.

\section*{Datasheet for \textsc{Infini-News}}

Following the framework proposed by \citet{Gebru2021}, we provide documentation for \textsc{Infini-News}.

\subsection*{Motivation}

\textbf{For what purpose was the dataset created?}
To provide researchers with efficient access to the Common Crawl News archive without requiring terabyte-scale local storage or complex WARC parsing infrastructure.
The dataset addresses the gap between the theoretical openness of CC-News and its practical inaccessibility due to engineering overhead.

\textbf{Who created the dataset and on behalf of which entity?}
Ruggero Marino Lazzaroni, Kirill Solovev 

\textbf{Who funded the creation of the dataset?}
Independent research

\subsection*{Composition}

\textbf{What do the instances represent?}
Each instance represents a single news article extracted from the CC-News WARC archive.

\textbf{How many instances are there in total?}
\qty{1.357}{\Bil} articles with non-empty extracted text.

\textbf{Does the dataset contain all possible instances or is it a sample?}
The corpus represents the CC-News WARC output from August 2016 through the latest available monthly snapshot.
Records lacking extractable body content were excluded during processing.

\textbf{What data does each instance consist of?}
Processed text (boilerplate removed) along with metadata: article title, publication date, source URL, three inferred language labels (GlotLID v3 over the first \num{4096} characters, lingua over articles below \num{1000} characters, and CommonLingua over title+description+first \num{1024} characters from a 334-language byte-level model), and inferred geographic origin (per-domain cascade over ccTLDs, Wikidata, curated reference lists, structural HTML signals, and the article-language corpus rule).

\textbf{Is there a label or target associated with each instance?}
No task-specific labels.
Language and geographic metadata are provided as enrichments, not ground-truth annotations.

\textbf{Is any information missing from individual instances?}
Geographic labels are unavailable for 41.9\% of registered domains, predominantly generic-TLD outlets for which neither Wikidata, the curated reference lists, nor any structural HTML rule applied.
Language identification may be unreliable for very short articles or mixed-language content.

\textbf{Are relationships between individual instances made explicit?}
No explicit relationships.
However, the preservation of duplicate content enables researchers to infer syndication relationships through n-gram matching.

\textbf{Are there recommended data splits?}
No predefined splits.
The temporal structure (daily resolution) allows researchers to define splits appropriate to their tasks.

\textbf{Are there any errors, sources of noise, or redundancies?}
The corpus intentionally preserves duplicate content to enable syndication analysis.
Extraction quality depends on \texttt{trafilatura}'s performance on individual DOM structures, which may vary across sources.

\textbf{Is the dataset self-contained?}
Yes.
The indexes reference records within the released corpus; no external fetching is required.

\textbf{Does the dataset contain data that might be considered confidential?}
No.
All content derives from publicly accessible news websites.

\textbf{Does the dataset contain data that might be offensive or cause anxiety?}
The corpus reflects global news coverage, which includes reporting on violent events, hate speech, and other sensitive material.

\textbf{Does the dataset relate to people?}
Yes, in the broad sense that it contains text written by journalists about events involving people.
However, it does not contain private communications or individually contributed data.

\textbf{Is it possible to identify individuals from the dataset?}
The dataset contains published news articles that may name public figures and private individuals mentioned in news coverage.
No additional personally identifiable information beyond the original published content is included.

\subsection*{Collection Process}

\textbf{How was the data acquired?}
Extracted from CC-News WARC files using \texttt{fastwarc} for parsing and \texttt{trafilatura} for content extraction.

\textbf{What mechanisms or procedures were used?}
Automated batch processing of monthly CC-News snapshots.
See Section~\ref{sec:preprocessing} for technical details.

\textbf{If the dataset is a sample, what was the sampling strategy?}
No sampling was applied.
We processed all available CC-News WARC files for the observation period.
Records were excluded only if content extraction failed.

\textbf{Who was involved in the data collection process?}
The authors.
No crowdworkers or manual annotation was involved.

\textbf{Over what timeframe was the data collected?}
The corpus covers content captured by Common Crawl from August 2016 onward, refreshed with each new monthly snapshot.
Note that CC-News crawl timestamps may not precisely match original publication dates.

\textbf{Were any ethical review processes conducted?}
N/A

\textbf{Did you collect the data from individuals directly?}
No.
The data was obtained from Common Crawl's public archive of news websites.

\textbf{Were individuals notified about the data collection?}
Not applicable.
The underlying data collection was performed by Common Crawl.
Our contribution is indexing and redistribution of processed content from this public archive.

\subsection*{Preprocessing/Cleaning/Labeling}

\textbf{Was any preprocessing/cleaning/labeling done?}
Yes.
HTML boilerplate removal via \texttt{trafilatura}; three parallel language classifiers: GlotLID v3 (\texttt{cis-lmu/glotlid}, fastText) on the first \num{4096} characters of the extracted text; lingua (high-accuracy mode) on articles shorter than \num{1000} characters; CommonLingua (\texttt{PleIAs/CommonLingua}, byte-level CNN+attention) on title+description+first \num{1024} characters of body text, run on GPU and written to a sidecar parquet joined back by WARC record id; and per-domain geographic inference via a confidence-tiered cascade over ccTLDs, Wikidata, curated reference lists, structural HTML evidence (JSON-LD publisher addresses, imprint-page parses, footer-copyright entities, audience-locale tags), and an article-language corpus rule.
The GlotLID column is left empty for articles below 50 characters; lingua and CommonLingua both label those when the input is non-empty.
No deduplication was performed.

\textbf{Was the raw data saved in addition to the preprocessed data?}
The original CC-News WARC files remain publicly available from Common Crawl.
We do not redistribute raw WARC data.

\textbf{Is the software used to preprocess/clean/label the data available?}
Yes.
The processing pipeline is released in the project repository (\url{https://codeberg.org/ksolovev/infini-news}) under the MIT License.

\subsection*{Uses}

\textbf{Has the dataset been used for any tasks already?}
Not at the time of submission.

\textbf{Is there a repository that links to papers or systems that use the dataset?}
We will maintain a list of known uses in the project repository.

\textbf{What (other) tasks could the dataset be used for?}
Longitudinal discourse analysis, agenda-setting and framing studies, content syndication detection, cross-platform information flow analysis, multilingual media comparisons, LLM training data contamination auditing, and retrieval-augmented generation pipelines.

\textbf{Is there anything about the composition or collection that might impact future uses?}
The intentional preservation of duplicates means that frequency counts reflect syndication patterns, not unique article counts.
Researchers requiring deduplicated corpora should apply post-hoc filtering.
The per-domain geographic attribution provides approximate labels, validated at 88.8\% precision on a 225-host held-out set, and should not be treated as ground truth.

\textbf{Are there tasks for which the dataset should not be used?}
The dataset should not be used for: (1) training language models intended to generate disinformation; (2) commercial redistribution without consideration of original publishers' rights; (3) surveillance of individual journalists or editorial staff.

\subsection*{Distribution}

\textbf{Will the dataset be distributed to third parties?}
Yes.
The dataset is intended for the research community.

\textbf{How will the dataset be distributed?}
The pre-processed corpus is available through a gated repository (Hugging Face) requiring institutional affiliation and stated research purpose.
The indexes and retrieval toolkit are publicly available via Codeberg (\url{https://codeberg.org/ksolovev/infini-news}).

\textbf{When will the dataset be distributed?}
Upon publication.

\textbf{Will the dataset be distributed under a copyright or IP license?}
The toolkit is released under the MIT License.
Index metadata is released under CC-BY 4.0.
The corpus text remains subject to the original publishers' terms; access is provided for non-commercial research purposes.
Terms of use prohibit redistribution and commercial exploitation.

\textbf{Have any third parties imposed IP-based or other restrictions?}
The underlying content is sourced from Common Crawl, which operates under its own terms of use.
Users should comply with applicable copyright law in their jurisdiction.

\textbf{Do any export controls or other regulatory restrictions apply?}
None known.

\subsection*{Maintenance}

\textbf{Who will be supporting/hosting/maintaining the dataset?}
The first author, Ruggero Marino Lazzaroni at the University of Graz

\textbf{How can the owner/curator/manager be contacted?}
By emailing \texttt{ruggero.lazzaroni@uni-graz.at}

\textbf{Is there an erratum?}
Not at time of release.
Errata will be documented in the project repository.

\textbf{Will the dataset be updated?}
Yes.
We commit to annual updates incorporating new CC-News snapshots.

\textbf{Will older versions continue to be supported/hosted/maintained?}
Yes.
The corpus is append-only: new monthly snapshots are added without modifying earlier data, so all previously released records remain accessible.

\textbf{If others want to extend/augment/build on the dataset, is there a mechanism for them to do so?}
Contributions and issue reports can be submitted via the project repository.
We welcome community contributions of additional metadata layers (e.g., topic annotations, sentiment scores) that comply with the project's licensing terms.

\end{document}